\documentclass{article}

\usepackage{acra}
\usepackage{graphicx}
\usepackage{multirow}
\usepackage[table,xcdraw]{xcolor}
\usepackage{adjustbox}
\usepackage{amsmath}
\usepackage[hyphens]{url}
\usepackage{subcaption}
\usepackage{hyperref}
\usepackage{siunitx}
\usepackage{color, colortbl} % for colours in tables
\definecolor{Gray}{gray}{0.9} % color for highlighting cells in table

\title{Racing Towards Reinforcement Learning based control of an Autonomous Formula SAE Car}
\author{
  Aakaash Salvaji, Harry Taylor, David Valencia, Trevor Gee, Henry Williams$^*$\\
  Centre for Automation and Robotic Engineering Science\\
  The University of Auckland, NZ\\
  \texttt{henry.williams@auckland.ac.nz}$^*$ \\
}

\begin{document}

\maketitle

\begin{abstract}
    With the rising popularity of autonomous navigation research, Formula Student (FS) events are introducing a Driverless Vehicle (DV) category to their event list. 
    This paper presents the initial investigation into utilising Deep Reinforcement Learning (RL) for end-to-end control of an autonomous FS race car for these competitions.
    We train two state-of-the-art RL algorithms in simulation on tracks analogous to the full-scale design on a Turtlebot2 platform. 
    The results demonstrate that our approach can successfully learn to race in simulation and then transfer to a real-world racetrack on the physical platform.
    Finally, we provide insights into the limitations of the presented approach and guidance into the future directions for applying RL toward full-scale autonomous FS racing.
\end{abstract}

\section{Introduction}
    Formula Student (FS) events are held worldwide for university teams to enter and race a custom-designed formula-style race car. 
    Most of these events are sponsored and regulated by the Society of Automotive Engineers (SAE) standards. 
    These events are labelled Formula: Society of Automotive Engineers (F:SAE). 
    As the popularity of Artificial Intelligence (AI) rises, the range of applications it can be applied to has also increased, including autonomous driving \cite{zhu2021deep,zhao2020sim}. 
    In 2017, Formula Student Germany (FSG) introduced a Driverless Vehicle (DV) category to their event list, marking it as the first FS competition to host such a class \cite{jun2018autonomous}.
    The Society of Automotive Engineers-Australasia (F:SAE-A) competition held in Australia each year is expected to introduce a driverless vehicle category soon.

    Autonomous navigation is a complex problem that requires a robot to solve the three fundamental problems of navigation: localisation, goal recognition, and path-planning \cite{gul2019comprehensive}. 
    If the environment map is available, a robot can create a global path and then apply a local planner to follow the path and avoid obstacles.
    However, prior knowledge of the environment is not always available in practice, and dynamic obstacles present a complex challenge for conventional planning algorithms \cite{mohanan2018survey}. 
    Ideally, a robotic navigation system should be able to adapt its path planning and behaviour to overcome various obstacles within an environment without the need for specialised planning approaches. 
    A specialised planning approach is designed for an environment or pre-programmed responses to states or objects. 
    For example, detecting doorways or chairs is not generally applicable to operating in outdoor environments. 

    A specialised path-planning approach to FSAE would require a complete mapping of all eventualities of these forms of states to responses for the robot to follow during racing.  
    Developing such a mapping is infeasible given the potentially infinite possible state-response pairs required to navigate during a race, especially with other cars on the track \cite{zhu2021deep,zhao2020sim}. 
    What is required is a means for the robot to be capable of autonomously learning and generalising these beneficial behaviours from its own experiences within an environment.

    A rapidly growing area of research seeks to solve several robotic control problems through Reinforcement Learning (RL). 
    RL derives a suitable control solution via direct interactions with the environment, enabling it to adapt to novel situations without human-engineered solutions. 
    This way of learning has made RL popular in the robotics community and other areas, emerging as a potential solution for increasing the adaptability of robotics systems in diverse scenarios \cite{ccalicsir2019model,liu2021policy,ramirez2021model,dewanto2020average,sato2019model}.

\begin{figure}[htb]
        \centering
        \includegraphics[width=0.8\linewidth]{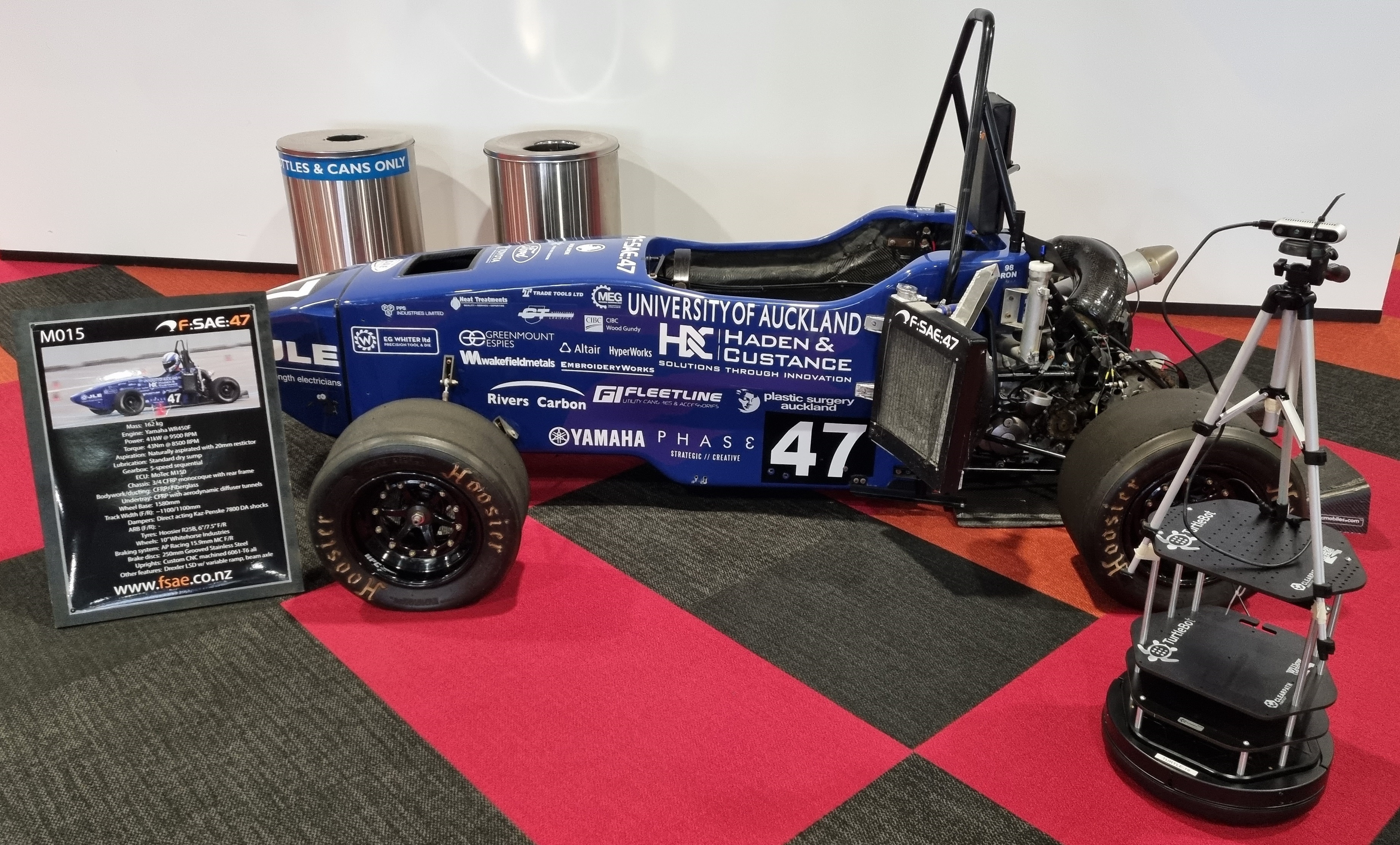}
        \caption{Turtlebot2 Robotic Platform (right) used for simulated and real-world training next to a full-scale FSAE car (left).}
        \label{fig:turtlbot2}
    \end{figure}

    This paper presents the initial investigation into utilising Deep Reinforcement Learning for end-to-end control of an F:SAE car.
    In place of a full-sized F:SAE race car, this work uses the Turtlebot2 robotic platform equipped with a Realsene D435 camera, shown in figure \ref{fig:turtlbot2}.
    Although not as fast as a racecar, it is a slow and steady platform with which we can reliably explore racing.
    We benchmark two state-of-the-art RL algorithms in continuous (TD3 \cite{fujimoto2018addressing}) and discrete (DQN \cite{van2016deep}) action spaces and provide insights into the challenges faced in applying RL to autonomous mobile navigation.
    Results are generated in simulated and real-world environments, and a detailed analysis is given.
    The code and resources to reproduce the results are provided via Github for others to use and apply for their research\footnote{\url{https://cares.blogs.auckland.ac.nz/research/reinforcementlearning/sae-car/}}.

\section{Background and Related Work}
    RL is an Artificial Intelligence (AI) branch that has recently gained tremendous research and industry impulse. 
    In contrast to unsupervised and supervised learning, RL does not need a large training dataset, an external teacher or pre-labelled data to learn a new task; instead, it derives the ground truth from an interaction with a surrounding environment. 
    As mention in \cite{sutton2018reinforcement} \textit{"Reinforcement learning is learning what to do"}. 
    This learning from experiences method resembles how humans and animals learn; even some studies reflect that the cerebellum may act as a reinforcement learning agent \cite{masettycerebellum}.

    A significant amount of research has been conducted in this area, with the performance of RL making significant strides in the last decade \cite{ccalicsir2019model,liu2021policy,ramirez2021model,dewanto2020average,sato2019model,arulkumaran2017deep}.
    Learning to play board games like Go on a professional level \cite{silver2016mastering} or reaching the human level playing video games like Atari \cite{mnih2013playing} are some examples of the success of RL.
    
    The specific algorithms explored in this work are the DQN \cite{mnih2013playing} and TD3 \cite{fujimoto2018addressing} networks. 
    DQN is a discrete value-based approach successfully used in several applications. 
    Although alternatives such as Double-DQN \cite{van2016deep} and Dueling-DQN \cite{wang2016dueling} exist, they do not always show significantly improved performance over the base DQN approach. 
    TD3 is a continuous policy-based approach that has shown relatively superior performance across a range of tasks \cite{fujimoto2018addressing}.
    These techniques provide a key insight into the two types of current RL approaches. 
    
    \subsection{RL for Mobile Robotic Control}
        Despite the success of using conventional approaches to tackle motion planning and control for mobile robot navigation, these approaches still require extensive engineering effort before they can reliably be deployed in the real world \cite{xiao2020motion,arulkumaran2017deep}. 
        As a way to potentially reduce this human engineering effort, many navigation methods based on machine learning have been recently proposed in the literature. 

        The surveys conducted in \cite{kiran2021deep,xiao2022motion} focus on Deep RL (DRL)  methods for autonomous driving. 
        They note that DRL has been successfully applied to controller optimisation, dynamic path planning, and trajectory optimisation. 
        It is discussed that DRL methods have the potential to be particularly effective in dynamic and unpredictable environments where traditional methods usually fail.
        Although the current DV challenge operates on a static track with a single car at a time, the long-term future may consider racing against other cars, which would introduce a complex navigation challenge. 
        
        Several challenges related to DRL methods are identified in \cite{niroui2019deep,kiran2021deep,xiao2022motion}, including validating performance, covering the simulation-reality gap, sample efficiency, designing practical reward functions and ensuring safety. 
        It is stressed that, like in \cite{arulkumaran2017brief}, most DRL methods presented are only  proven in simulation and have not been tested in the more complex physical world.
        DRL models are generally trained in simulation to reduce training time, and effort \cite{zhang2018robot,Williams2019}. 
        As discussed in \cite{kiran2021deep}, this creates a ``simulation to reality gap'' where the trained model does not perform as well in real life as it does in simulation. 
        
        Several approaches are applied to improve the transferability of the trained model to the real world.
        These include domain randomisation (randomly altering parameters of the simulation environment such as sensor noise), transfer learning (training a model trained in a source environment in a new target environment) and domain adaptation (a sub-classification of transfer learning where the source is a simulation and the target is the physical world).
        These approaches have reduced the simulation-to-reality gap in some applications, but the challenge is still unresolved.

        % The work in \cite{niroui2019deep} utilises DRL Robot for search and rescue applications to enable the robot to navigate unknown environments.
        % They developed a unique approach that combines an A3C \cite{mnih2016asynchronous} network with frontier exploration to learn an efficient exploration strategy based on high-dimensional robot states.
        % Like most work in this area, the training was conducted in the Stage simulator, but the experiments were run on a real-world Turtlebot2 platform. 
        % The training translated to the real world, showing improved performance for exploring the environment in the least amount of time.
        % The work, however, was only evaluated in a static environment. 

        The work in \cite{li2021behavior} presents a behaviour-based mobile robot navigation method with DRL.
        Utilising DQN \cite{van2016deep} and PPO \cite{schulman2017proximal}, the work trains four simulated robot platforms to coordinate their movements to swap places in a static obstacle-less environment.
        The simulated results show that the robots can manoeuvre around each other without collisions or knowledge shared between the robotic platforms. 
        Two Pioneer 3-AT robots are shown successfully navigating in a real-world dynamic environment with humans walking past the robots without collisions using only the simulated training.
        The work only uses a 2D LIDAR sensor, which would not provide enough information in more complex environments. 
    
        The work in \cite{remonda2021formula} utilises DRL for autonomous racing using telemetry data on the TORCS race car simulator.
        Training variations of the DDPG \cite{lillicrap2015continuous} the race car was capable of learning to navigate the race tracks and was even able to generalise to unseen tracks post-training. 
        Although this work was conducted entirely in simulation, this work does show the promise of using DRL to control a race car on known and unknown tracks. 

        The only other work we are aware of that is seeking to apply RL to the FSAE challenge is work conducted by \cite{zadok2019explorations}. This work uses an imitation learning algorithm with a modified PilotNet artificial neural network (ANN) for end-to-end autonomous driving of an FSAE race car. 
        To train the model, data was gathered by human drivers in model F:SAE vehicles using the photo-realistic AirSim simulation software. 
        This method is limited because the autonomous vehicle can never outperform human drivers and can not learn novel behaviours outside human-generated data. 
        However, the work was impressively demonstrated on a full-sized autonomous FSAE race car with training only conducted in simulation.

\section{Methods}\label{sec:methods}
    We evaluated the discrete DQN and continuous TD3 approaches for learning to control the Turtlebot2 around a simulated and real-world race track.
    To produce a fair and consistent evaluation, each algorithm was trained under the same conditions and was compared on the same race tracks post-training. 

    \subsection{Autonomous FSAE Task}
\begin{figure}[htb]
            \centering
            \includegraphics[width=0.65\linewidth]{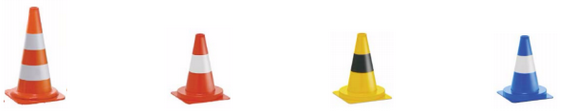}
            \caption{Examples of the cones used for FSG DV events that define the race track's path.}
            \label{fig:cones}
        \end{figure}

        F:SAE-A currently has no DV class, but to allow the preemptive development of an autonomous F:SAE vehicle, we can look to the rule set used at the FSG events and expect similar guidelines to be implemented in a future F:SAE-A DV class. 
        The FSG 2022 rule book \cite{avrules2022} and the competition handbook \cite{avhandbook2022} state that the DV event tracks are defined using the cones shown in figure \ref{fig:cones}. 
        It should also be noted that according to \cite{avhandbook2022}, no official maps of the tracks are provided to contestants.
        This means the autonomous system can not be trained to a specific track in advance and will need to be able to handle a novel track to compete on the day.
        The current DV class also only considers a single car racing on the track at a given time, racing only against the time to complete the course. 
    
\begin{figure}[htb]
            \centering
             \begin{subfigure}[]{0.49\linewidth}
                 \centering
                 \includegraphics[height=3.5cm]{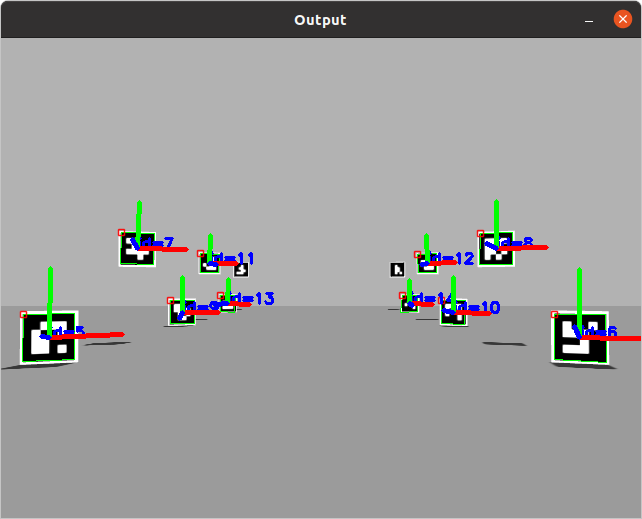}
                 \caption{Simulated Straight Track}
                 \label{fig:sim-straight}
             \end{subfigure}
             \hfill
             \begin{subfigure}[]{0.49\linewidth}
                 \centering
                 \includegraphics[height=3.5cm]{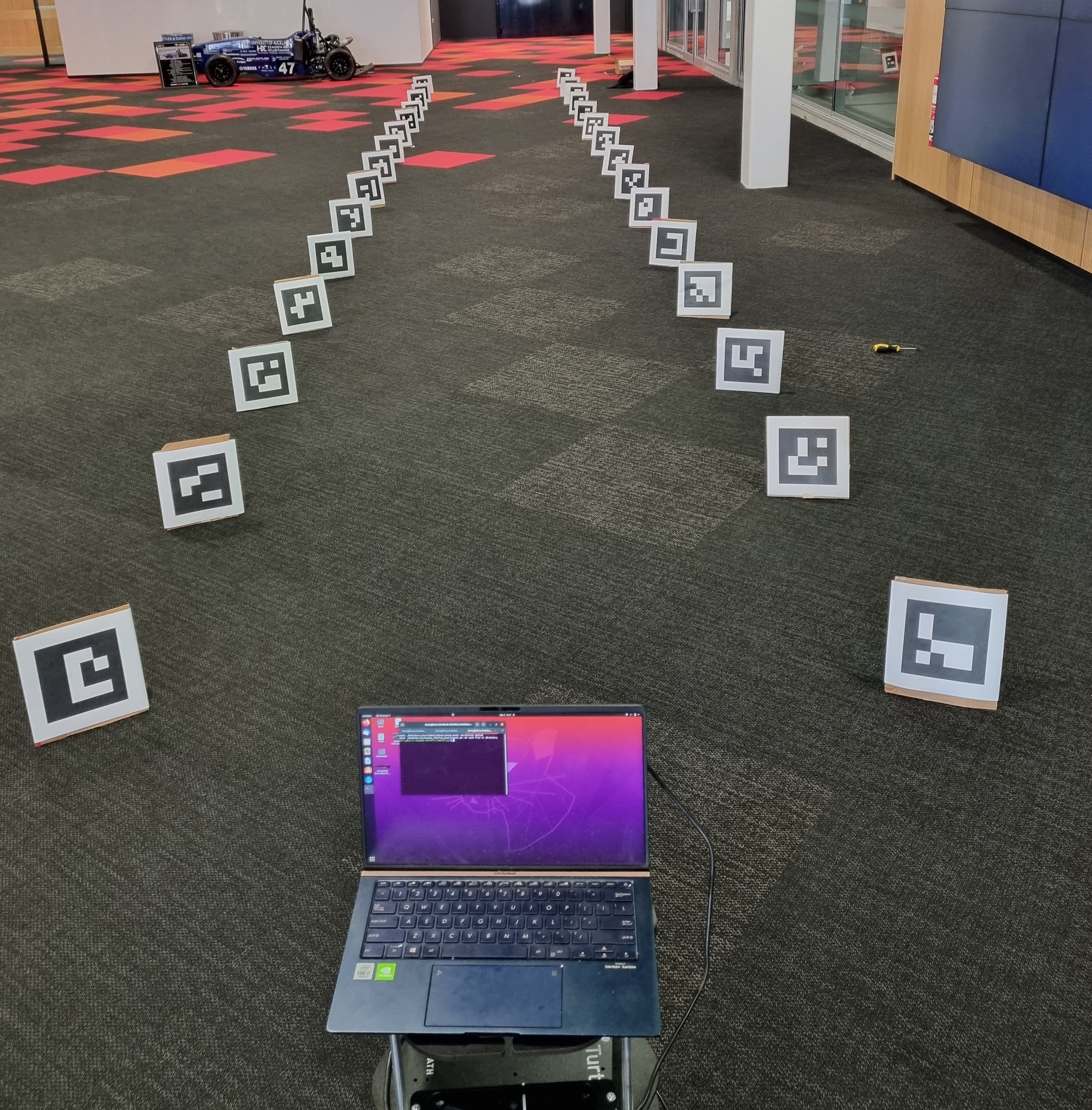}
                 \caption{Real World Straight Track}
                 \label{fig:real-straight}
             \end{subfigure}
             \caption{Simulated and real-world race track utilising the Arcuo markers in place of cones}
                \label{fig:track-example}
        \end{figure}

        The work presented in this paper replaces the cones with ArUco markers \cite{garrido2014automatic} to constrain the computer vision challenge and focus on the control problem. 
        This representation will also translate between simulation and the real world as it does not rely on specific visual information.
        An example of the virtual and real-world track created using the ArUco markers is shown in figure \ref{fig:track-example}. The markers were placed two units across and 1 unit apart along the track, where 1 unit in simulation translates to \SI{0.85}{\meter} in the real world track.
        
        The objective of the Turtlebot2 platform is to race down the centre-line of the race track. 
        The centre-line is used as defining an effective racing line reward function is beyond the scope of this work.
        The tracks are generated with odd ArUco ID markers on the left and even on the right to assist automatic detection of the track's centre line for generating a reward.
        Future work that reliably detects the cones can replace the ArUco detection approach.

    \subsection{State and Action Space}
        The state space used to train both models was the positions of the six closest Aruco markers detected by the RealSense D435 camera connected to the Turtlebot2. 
        The position information was formatted as the $x$ (lateral distance) and $z$ (forward distance) position of the markers relative to the Turtlebot2 in ascending order of $z$. 
        To simplify the action space, both models only controlled the angular velocity of the robot while the linear velocity was constant at \SI{0.5}{\meter\per\second}. 
        The linear and angular velocities were sent to the Turtlebot2 via a ROS publisher using a Twist message. 

        The DQN model's discrete action space was represented by a binary positive or negative fixed rotation speed (set to $\pm$\SI{0.2}{\radian\per\second}). 
        The TD3 model's continuous action space was represented by a single float value normalized between $-1$ and $1$ and then scaled by \SI{0.4}{\radian\per\second}, allowing continuous rotational velocity values from \SIrange{-0.4}{0.4}{\radian\per\second}.
        The TD3 rotation potential speed was set higher as the continuous space allowed for more refined control while turning, whereas a higher speed on DQN caused the Turtlebot2 to overshoot and go off track consistently. 
        
        The network architecture for the DQN network consists of two fully connected layers of 1024 and 512 nodes, with \textit{ReLU} activation functions for both layers and a \textit{Linear} activation function for the output layer. The DQN network uses a batch size of 64 and a learning rate of $1e^{-4}$. 

        The TD3 network consists of an actor-network and two critic-networks. 
        All three networks have three fully connected layers with 64, 64 and 32 nodes. 
        The three layers of the actor-network have a \textit{ReLU} activation function with a \textit{Tanh} output activation function. 
        The three layers of the critic-networks have \textit{ReLU} activation functions with a \textit{Linear} output activation function. The TD3 network uses a batch size of 32 with a learning rate of $1e^{-4}$ for the actor-network and a learning rate of $1e^{-3}$ for the critic-network.
        Full details of the models and their structures can be found at the Github link.

    \subsection{Training Procedure}
        The robot was trained by racing it through a series of short simulated track segments created by the ArUco markers, as shown in figure \ref{fig:track-example}. 
        Track segments were generated for each episode with a fixed inner turn radius of 10 units and a turn angle randomly set to a multiple of 15 degrees between -90$^{\circ}$ to 90$^{\circ}$ (i.e., -90$^{\circ}$, -75$^{\circ}$, -60$^{\circ}$ …. 60$^{\circ}$, 75$^{\circ}$, 90$^{\circ}$) to allow the model to train on a wide variety of turn angles. For a basic straight track segment, the markers are separated by 1 unit along the track and 2 units apart, with 20$^{\circ}$ internal rotation towards the centre of the track. The track width was kept constant at 2 units wide for turns, but the consecutive marker separation depended on turn angle. ArUco markers were placed along the curve with equidistant spacing. Alternating marker pairs were elevated to reduce the visual obstruction of markers. This does not affect the state space as the markers' elevation was irrelevant. A similar effect to reduce obstruction can be achieved in the real world by elevating the robot's camera.

        The robot was initially trained on a single segment (using 15 pairs of ArUco markers) of either a straight, a 90$^{\circ}$ left turn, or a 90$^{\circ}$ right turn. 
        The model showed semi-consistent completion of navigating the track segment after 5000 episodes (43 of the last 100 training episodes). 
        Although this initial training was only performed on straights and 90$^{\circ}$ turns, the model showed some success in navigating 30$^{\circ}$ and 60$^{\circ}$ turns, but not consistently enough to suggest it had completed the task. 
        Furthermore, when multiple segments were chained together to test the model, it consistently failed after the first segment if the first segment was a turn, leading us to believe that the model had not learned to re-centre itself effectively when coming out of a turn. 
        To address this problem, we increased the length of the training track to two randomly connected track segments. This allowed for a large variety (13$^2$ = 169) of track variations to train on and let the robot learn to navigate multiple turns at various angles.
    
        Given that the training is conducted purely in simulation, there is a risk of running into the simulation-to-reality gap. 
        Specifically, the precise placement of the ArUco markers in the simulation may not be easily replicated when testing in the real world. 
        To minimise this risk, artificial noise was incorporated into the position of the ArUco markers along the track segments.
        Each ArUco marker is randomly displaced by $\mp0.05$ units and rotated by $\mp15^{\circ}$ after being placed along the track segment.
        The models are trained with and without noise on the tracks to evaluate if this impacts transferring the learned policy to the real world.

        Each model was trained in simulation for 5000 episodes (track segments) with an allowance of 2000 steps (actions) per episode. Each model state was saved separately every 500 episodes.
        The Turtlebot2 was given Twist commands at a rate of 5 actions per second. 
        The episode was terminated either upon detection of the '0' ID Aruco marker used to indicate the end of the track, the Turtlebot2 got too close to any Aruco marker ($\leq\SI{0.5}{units}$),  or if at least one odd and one even marker (required for reward feedback) was not visible for ten sequential actions.
        
    \subsection{Reward Function}
\begin{figure}[htb]
            \centering
             \begin{subfigure}[]{0.49\linewidth}
                 \centering
                 \includegraphics[width=\linewidth]{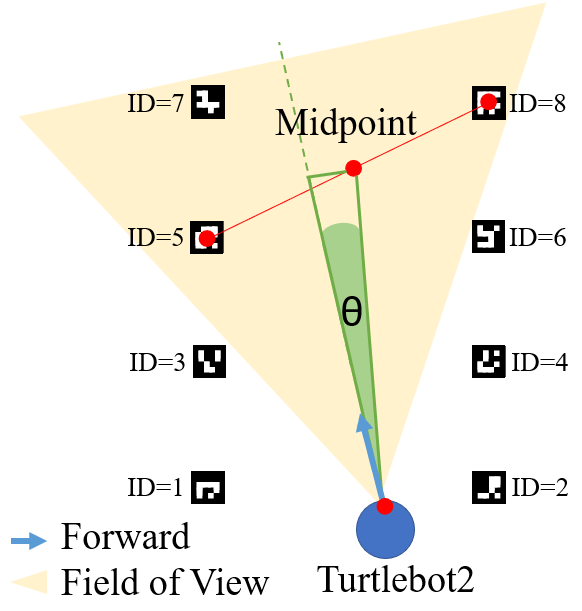}
                 \caption{Diagram showing how the angle $\theta$ is calculated.}
                 \label{fig:reward-function-diagram}
             \end{subfigure}
             \hfill
             \begin{subfigure}[]{0.49\linewidth}
                 \centering
                 \includegraphics[width=\linewidth]{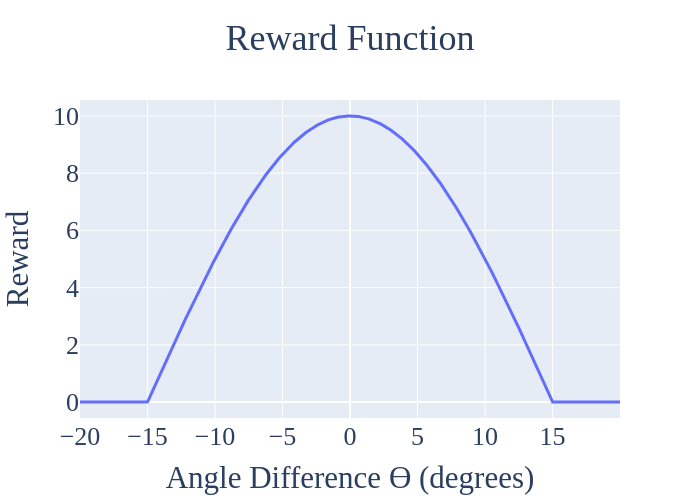}
                 \caption{Plot of reward function relative to $\theta$}
                 \label{fig:reward-function-plot}
             \end{subfigure}
             \caption{Visualisations of the reward function.}
                \label{fig:reward-function}
        \end{figure}

         A dense reward function is used to provide the system with consistent feedback, with each action taken by the robot receiving a reward.  The reward function is based on the angle difference between the current direction the Turtlebot2 faces and the midpoint of the closest pair of markers. The reward is designed to keep the Turtlebot2 close to the centreline of the track with higher rewards for being aligned with the midpoint. The cosine function is ideal, as the output decreases as the angle difference increases in either direction and is highest when the angle is 0$^{\circ}$. The reward function is defined as:
        \[
        Reward = S\times \cos(\theta \times A), Reward\geq 0
        \]
        Where $\theta$ is the angle difference between the robot's forward direction and direction towards the midpoint of the closest pair as shown in figure \ref{fig:reward-function-diagram}.
        
        For training, the reward scalar ($S$) was set to 10, the angle drop-off ($A$) was set to 6, and clipped negative reward values to 0. This function provides a maximum reward per step of 10 if the angle difference is zero and zero rewards when the angle is equal to or above $\mp15^{\circ}$ as shown in figure \ref{fig:reward-function-plot}. The reward scalar was chosen arbitrarily so that the max reward possible at any given timestep is 10. The angle drop-off value was chosen based on empirical testing that showed that an angle difference greater than $\mp15^{\circ}$ allowed the robot to learn unwanted behaviours such as over-steering and driving off the track.
        
        A marker “pair” was defined as one odd ID ArUco marker (left side) and one even ID ArUco marker (right side), e.g. 5\&8. The positions of the lowest pair (lowest odd and even) are averaged to find the midpoint, giving an estimate of the centre-line of the track and the direction the robot should advance. Although using consecutive marker IDs (i.e., 1\&2, 3\&4, etc.) would give a more accurate estimate of the centre-line and trajectory of the immediate track, it was found that occasional obstructions of markers terminated training episodes early, as no suitable pair could be found. Although sacrificing some accuracy in the estimated centre-line, the chosen “pair” definition proved more reliable in the case of obstructions and detection issues.

    \subsection{Video Demonstration}
         A video demonstrating the learning process and results are shared \href{https://cares.blogs.auckland.ac.nz/research/reinforcementlearning/sae-car/}{here}\footnote{\url{https://cares.blogs.auckland.ac.nz/research/reinforcementlearning/sae-car/}}. The video shows each model's performance on a two-segment track consisting of a 90$^{\circ}$ left turn followed by a 90$^{\circ}$ right turn, using the checkpoint models saved at 500 episodes, 2500 episodes and 5000 episodes. The video also contains examples showing how real-world and oval track testing was conducted.
    
\section{Results}
    DQN and TD3 were trained in simulation using the approach described in Section \ref{sec:methods}.
    The learned racing performance of DQN and TD3 is evaluated both in simulation and in the real world. Performance was quantified using two metrics; 1) The number of times the robot reached the end of the track (Finish \%), and 2) The average distance completed along the total length of the track (AVG Distance \%).

    \subsection{Simulation Training Results}
\begin{figure}[htb]
            \centering
            \includegraphics[width=\linewidth]{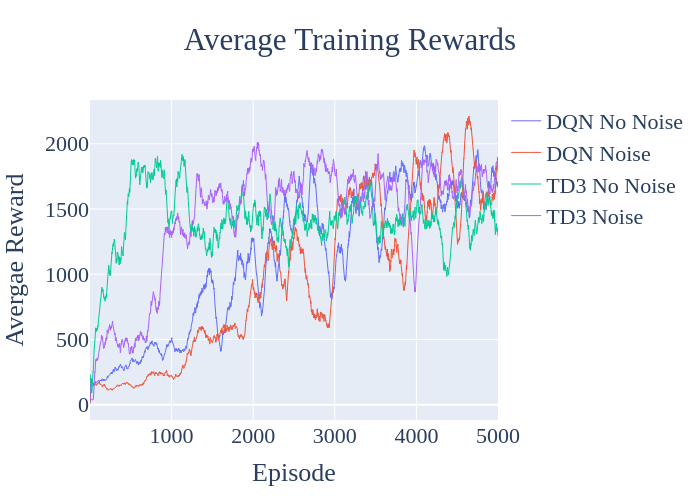}
            \caption{Rewards graph from training models for 5000 episodes.}
            \label{fig:rewards-graph}
        \end{figure}

        Each model was trained for 5000 episodes. The training process results for DQN and TD3 are shown in figure \ref{fig:rewards-graph}. Both algorithms show increasing rewards, confirming that they can learn the task. TD3 begins to plateau within 2000 episodes, whereas DQN’s average reward is still rising at 5000 episodes. Training DQN further may improve performance, but the models were not trained further due to time constraints.

    \subsection{Simulation Track Segment Results}
        The trained models were tested the same way as training by connecting two track segments. As mentioned above, 13 track segment variations can be connected into two segment sections in 169 combinations (13$^2$=169). Each model was tested for 169 episodes, once on each possible combination. A summary of the results of this testing can be seen in table \ref{tab:simulated-segment-results}. Although the results show that both TD3 models performed similarly, analysis of individual episode completions shows that the TD3 model trained without noise performs better on sharp left turns than on sharp right turns, whereas the TD3 model trained with noise shows the opposite, performing better on sharp right turns than lefts. 
        
        % TABLE FOR SIMULATION TRACK SEGMENT TESTING
        \begin{table}[htb]
        \caption{Success rate on simulated racing segments} 
        \centering
        \begin{adjustbox}{width=\linewidth}
        \begin{tabular}{|c|l|l|l|}
        \hline
        \multirow{2}{*}{\textbf{MODEL}}                           & \multirow{2}{*}{\textbf{TRACK}} & \multirow{2}{*}{\textbf{Finish \%}} & \multirow{2}{*}{\textbf{AVG Distance \%}} \\
                                                                  &                                 &                                     &                                             \\ \hline
        \multirow{2}{*}{\textbf{DQN}}                             & \textbf{No Noise}               & 37.28                               & 69.68                                       \\ \cline{2-4} 
                                                                  & \textbf{Noise}                  & 33.14                               & 61.68                                       \\ \hline
        \multicolumn{1}{|l|}{\multirow{2}{*}{\textbf{DQN Noise}}} & \textbf{No Noise}               & 59.76                               & 76.14                                       \\ \cline{2-4} 
        \multicolumn{1}{|l|}{}                                    & \textbf{Noise}                  & 60.95                               & 77.2                                        \\ \hline
        \multirow{2}{*}{\textbf{TD3}}                             & \textbf{No Noise}               & 68.64                               & 81.38                                       \\ \cline{2-4} 
                                                                  & \textbf{Noise}                  & 72.19                               & 84.32                                       \\ \hline
        \multicolumn{1}{|l|}{\multirow{2}{*}{\textbf{TD3 Noise}}} & \textbf{No Noise}               & 70.41                               & 79.71                                       \\ \cline{2-4} 
        \multicolumn{1}{|l|}{}                                    & \textbf{Noise}                  & 61.54                               & 77.08                                       \\ \hline
        \end{tabular}
        \end{adjustbox}
        \label{tab:simulated-segment-results}
        \end{table}
        
    \subsection{Real-World Track Segments}
\begin{figure}[htb]
            \centering
             \begin{subfigure}[]{0.49\linewidth}
                 \centering
                 \includegraphics[width=\linewidth]{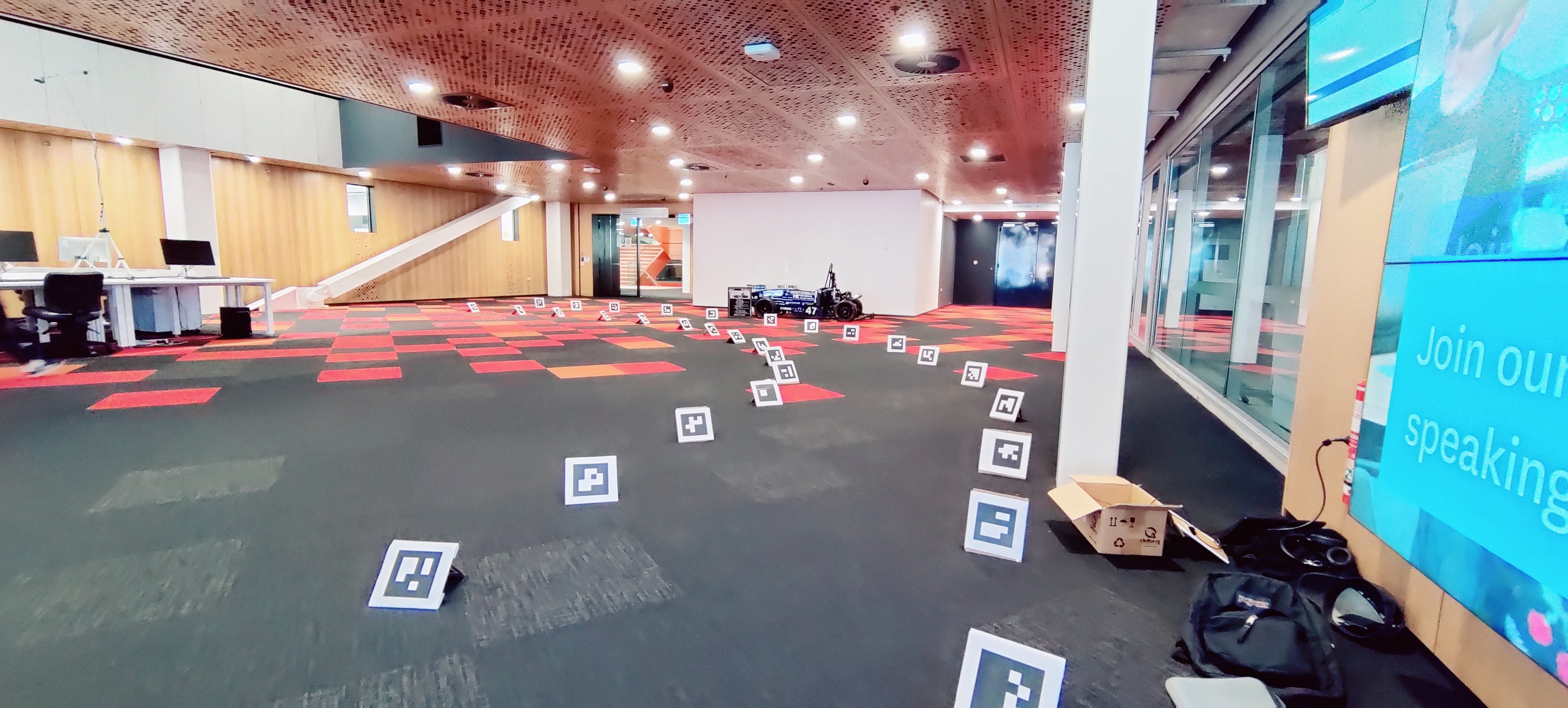}
                 \caption{Real 90$^{\circ}$ Left Track}
                 \label{fig:left-track}
             \end{subfigure}
             \hfill
             \begin{subfigure}[]{0.49\linewidth}
                 \centering
                 \includegraphics[width=\linewidth]{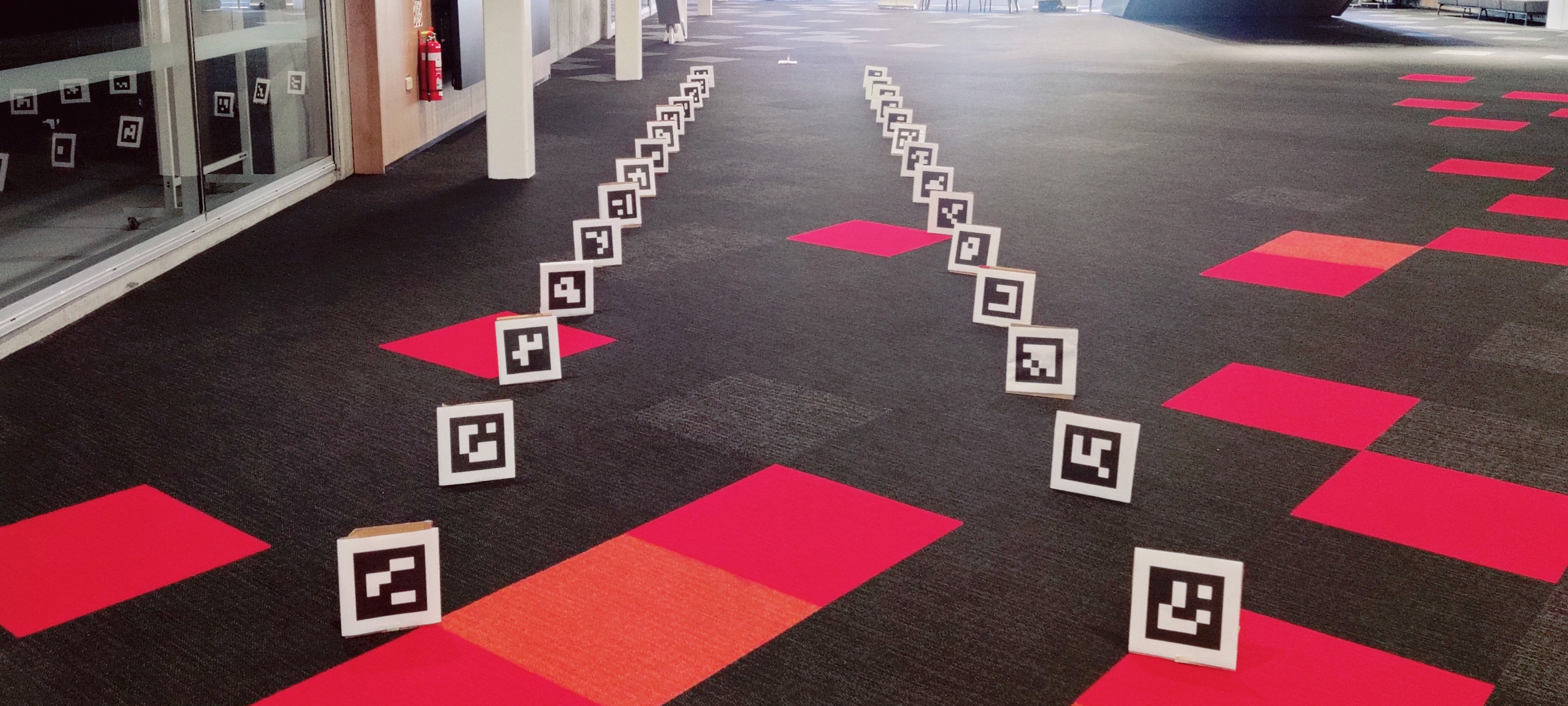}
                 \caption{Real Straight Track}
                 \label{fig:straight-track}
             \end{subfigure}
             \hfill
              \begin{subfigure}[]{0.49\linewidth}
                 \centering
                 \includegraphics[width=\linewidth]{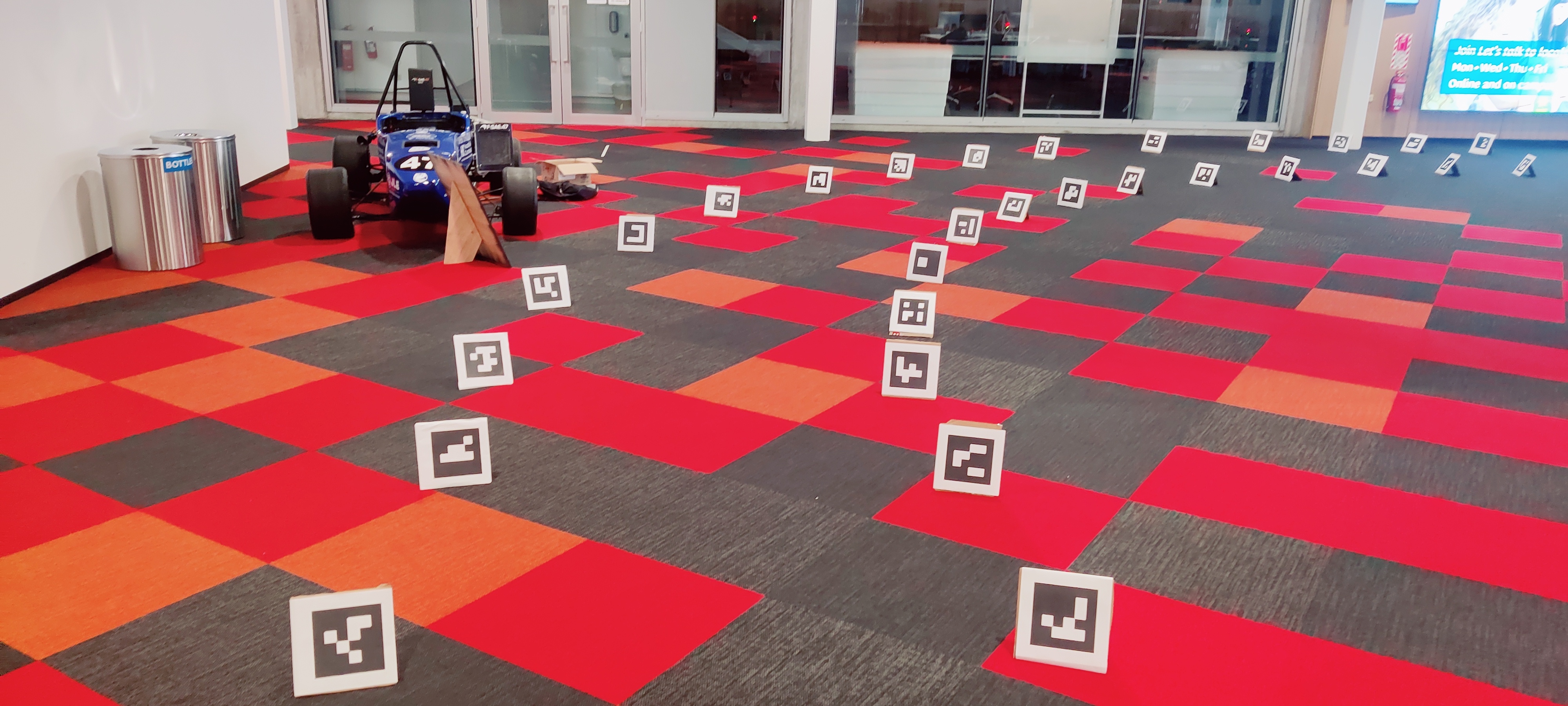}
                 \caption{Real 90$^{\circ}$ Right Track}
                 \label{fig:right-track}
             \end{subfigure}
             \caption{Real-world race track utilising the Arcuo markers in place of cones}
                \label{fig:real-track-examples}
        \end{figure}

        Over a week, the trained models were evaluated in the real world on straight, left turn, and right turn track segments using physical ArUco markers.
        These segments were set up to match the simulated training scenarios as closely as possible, albeit with only the first half of the segment due to space restrictions.
        The track segments consisted of 15 ArUco marker pairs, similar to the simulation. The ArUco markers were spaced as accurately as possible to replicate the spacing in the simulation, where 1 Unit in simulation translated to \SI{0.85}{\meter} in the real world.
        An example of each of the left turn, straight and right turn tracks are shown in figure \ref{fig:real-track-examples}. 
     
        The robot was given 3 starting positions in the simulation testing, in the centre of the track and a \SI{0.425}{\meter} displacement to the left and the right. Each model was run 30 times along the 3 tracks (10 times per starting position), totalling 90 runs per model, with the percentage of track segment completion rate measured for each run. The results of the real-world testing are shown in table \ref{tab:real-segment-results}.
        
        The results show that the TD3 models performed better than the DQN models in the real world. As expected, the DQN model trained with noise performed slightly better than the model trained without noise. However, the TD3 model trained without noise outperforms the noisy TD3 model overall. Although it was expected that the noisy TD3 would perform better on the right turn, the results show that both TD3 models perform better on left turns than on the right turns, contradicting the simulated track segment testing results. Overall, the simulation training transferred to the real-world track showed success rates between 60\% and 85\% average completion. There is room for improvement in completing the full track reliably, but the results are positive. 
        
        % REAL-WORLD RESULTS        
        \begin{table*}[htb]
        \caption{Success rate on real-world racing segments} 
        \begin{adjustbox}{width=\textwidth}
        \begin{tabular}{|
        >{\columncolor[HTML]{FFFFFF}}c |
        >{\columncolor[HTML]{FFFFFF}}l 
        >{\columncolor[HTML]{FFFFFF}}l |
        >{\columncolor[HTML]{FFFFFF}}l 
        >{\columncolor[HTML]{FFFFFF}}l |
        >{\columncolor[HTML]{FFFFFF}}l 
        >{\columncolor[HTML]{FFFFFF}}l |
        >{\columncolor[HTML]{DAE8FC}}l 
        >{\columncolor[HTML]{DAE8FC}}l |}
        \hline
        \textbf{}          & \multicolumn{2}{c|}{\cellcolor[HTML]{FFFFFF}\textbf{Left}}                               & \multicolumn{2}{c|}{\cellcolor[HTML]{FFFFFF}\textbf{Straight}}                           & \multicolumn{2}{c|}{\cellcolor[HTML]{FFFFFF}\textbf{Right}}                              & \multicolumn{2}{c|}{\cellcolor[HTML]{DAE8FC}\textbf{Total}}                             \\ \hline
        \textbf{Model}     & \multicolumn{1}{l|}{\cellcolor[HTML]{FFFFFF}\textbf{Finish \%}} & \textbf{AVG Distance \%} & \multicolumn{1}{l|}{\cellcolor[HTML]{FFFFFF}\textbf{Finish \%}} & \textbf{AVG Distance \%} & \multicolumn{1}{l|}{\cellcolor[HTML]{FFFFFF}\textbf{Finish \%}} & \textbf{AVG Distance \%} & \multicolumn{1}{l|}{\cellcolor[HTML]{DAE8FC}\textbf{Final \%}} & \textbf{AVG Distance \%} \\ \hline
        \textbf{DQN}       & \multicolumn{1}{l|}{\cellcolor[HTML]{FFFFFF}3.33}               & 55.33                  & \multicolumn{1}{l|}{\cellcolor[HTML]{FFFFFF}56.67}              & 87.56                  & \multicolumn{1}{l|}{\cellcolor[HTML]{FFFFFF}0}                  & 51.33                  & \multicolumn{1}{l|}{\cellcolor[HTML]{DAE8FC}20}                & 64.74                  \\ \hline
        \textbf{DQN Noise} & \multicolumn{1}{l|}{\cellcolor[HTML]{FFFFFF}10}                 & 56.44                  & \multicolumn{1}{l|}{\cellcolor[HTML]{FFFFFF}63.33}              & 88.22                  & \multicolumn{1}{l|}{\cellcolor[HTML]{FFFFFF}3.33}               & 54.22                  & \multicolumn{1}{l|}{\cellcolor[HTML]{DAE8FC}25.55}             & 66.29                  \\ \hline
        \textbf{TD3}       & \multicolumn{1}{l|}{\cellcolor[HTML]{FFFFFF}\textbf{60}}        & \textbf{90.88}         & \multicolumn{1}{l|}{\cellcolor[HTML]{FFFFFF}76.67}              & 88.89                  & \multicolumn{1}{l|}{\cellcolor[HTML]{FFFFFF}\textbf{30}}        & \textbf{64.44}         & \multicolumn{1}{l|}{\cellcolor[HTML]{DAE8FC}\textbf{55.56}}    & \textbf{81.40}         \\ \hline
        \textbf{TD3 Noise} & \multicolumn{1}{l|}{\cellcolor[HTML]{FFFFFF}33.33}              & 80                     & \multicolumn{1}{l|}{\cellcolor[HTML]{FFFFFF}\textbf{80}}        & \textbf{95.78}         & \multicolumn{1}{l|}{\cellcolor[HTML]{FFFFFF}13.33}              & 58.22                  & \multicolumn{1}{l|}{\cellcolor[HTML]{DAE8FC}42.22}             & 78                     \\ \hline
        \end{tabular}
        \end{adjustbox}
        \label{tab:real-segment-results}
        \end{table*}
    
    \subsection{Oval Race Track Simulation}
\begin{figure}[htb]
            \centering
             \begin{subfigure}[]{0.49\linewidth}
                 \centering
                 \includegraphics[width=\linewidth]{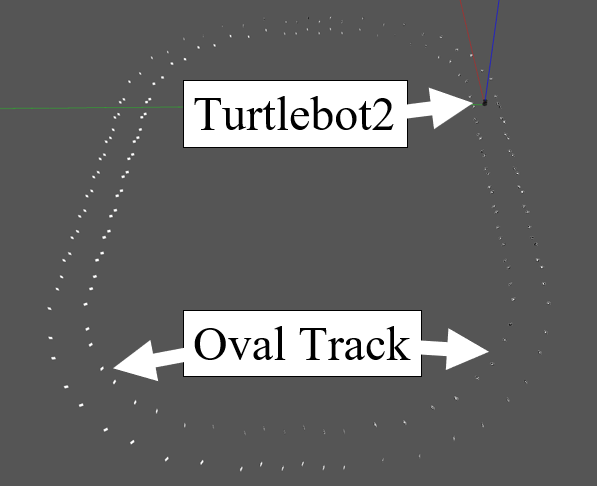}
                 \caption{The oval test track used in the simulation.}
                 \label{fig:oval-track-a}
             \end{subfigure}
             \hfill
             \begin{subfigure}[]{0.49\linewidth}
                 \centering
                 \includegraphics[width=\linewidth]{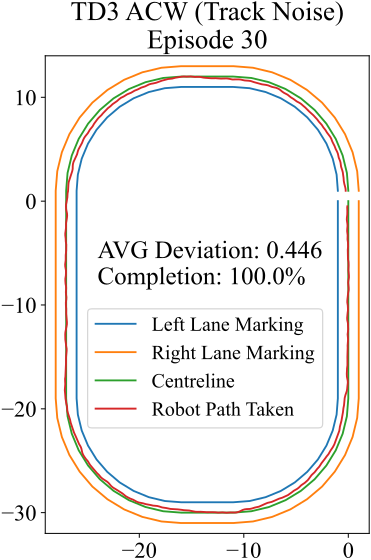}
                 \caption{Path of the Turtlebot2 using TD3 to race around oval track.}
                 \label{fig:results-example}
             \end{subfigure}
             \caption{Oval Track used for testing the racing performance.}
                \label{fig:oval-track}
        \end{figure}

        An oval track consisting of six track segments (90 Arcuo marker pairs), shown in figure \ref{fig:oval-track}, was set up to evaluate the racing performance of the trained models on a full-sized track. The track was generated with and without random noise. 
        % Loading all six segments at the same time, caused significant lag and performance issues due to large quantities of models being loaded. The track generation was changed to render only 2 segments at a time (current \& next), with the following segment being loaded when the Turtlebot2 crosses the boundary between two segments. 
        
        Each model, with and without noise, was used to race the Turtlebot2 platform around the oval track (clockwise and anti-clockwise). 
        The robot started at three start locations: in the centre, to the left and right side of the track.
        Each start location was run 30 times for 90 runs around the oval race track in each clockwise and anti-clockwise direction.
        Figure \ref{fig:results-example} shows an example of these runs for the TD3 model. The orange and blue lines represent the lane markings of the track, the green line represents the centre line, and the red line shows the path the robot took around the track.
        
        The number of times the model finished the track and the average distance completed were captured to measure the performance of each model and are presented in table \ref{tab:oval-track-results}.
        The results show no significant difference in performance between clockwise (turning right) and anticlockwise (turning left) runs for both DQN models. The TD3 models again performed better and more consistently than the DQN models. Like the track segment testing results, the TD3 model trained without noise showed significantly better performance turning left, and the TD3 model trained with noise performed better turning right. 
        Given the prior results both in simulation and the real world, the total success rate of the models on the oval track was lower than expected.
        All the models struggled to complete the full oval tracks, with an average completion rate below 60\%.
        The only pattern was that most episodes terminated while the robot was on a turn, indicating that the models had no problem navigating the straight portions.
    
        \begin{table*}[htb]
        \caption{Racing performance of the trained models on the simulated oval track}
        \begin{adjustbox}{width=\textwidth}
        \begin{tabular}{|
        >{\columncolor[HTML]{FFFFFF}}c |
        >{\columncolor[HTML]{FFFFFF}}l |
        >{\columncolor[HTML]{FFFFFF}}l 
        >{\columncolor[HTML]{FFFFFF}}l 
        >{\columncolor[HTML]{FFFFFF}}l 
        >{\columncolor[HTML]{FFFFFF}}l 
        >{\columncolor[HTML]{FFFFFF}}l |
        >{\columncolor[HTML]{FFFFFF}}l 
        >{\columncolor[HTML]{FFFFFF}}l 
        >{\columncolor[HTML]{FFFFFF}}l 
        >{\columncolor[HTML]{FFFFFF}}l 
        >{\columncolor[HTML]{FFFFFF}}l |
        >{\columncolor[HTML]{DAE8FC}}l 
        >{\columncolor[HTML]{DAE8FC}}l |}
        \hline
        \cellcolor[HTML]{FFFFFF}                                     & \cellcolor[HTML]{FFFFFF}                                 & \multicolumn{5}{l|}{\cellcolor[HTML]{FFFFFF}\textbf{Clockwise}}                                                                                                                                                                                                            & \multicolumn{5}{l|}{\cellcolor[HTML]{FFFFFF}\textbf{Anti-Clockwise}}                                                                                                                                                                                                       & \multicolumn{2}{l|}{\cellcolor[HTML]{DAE8FC}\textbf{Total}}                                  \\ \cline{3-14} 
        \multirow{-2}{*}{\cellcolor[HTML]{FFFFFF}\textbf{MODEL}}     & \multirow{-2}{*}{\cellcolor[HTML]{FFFFFF}\textbf{TRACK}} & \multicolumn{1}{l|}{\cellcolor[HTML]{FFFFFF}\textbf{L}} & \multicolumn{1}{l|}{\cellcolor[HTML]{FFFFFF}\textbf{C}} & \multicolumn{1}{l|}{\cellcolor[HTML]{FFFFFF}\textbf{R}} & \multicolumn{1}{l|}{\cellcolor[HTML]{FFFFFF}\textbf{Finish \%}} & \textbf{AVG Distance \%} & \multicolumn{1}{l|}{\cellcolor[HTML]{FFFFFF}\textbf{L}} & \multicolumn{1}{l|}{\cellcolor[HTML]{FFFFFF}\textbf{C}} & \multicolumn{1}{l|}{\cellcolor[HTML]{FFFFFF}\textbf{R}} & \multicolumn{1}{l|}{\cellcolor[HTML]{FFFFFF}\textbf{Finish \%}} & \textbf{AVG Distance \%} & \multicolumn{1}{l|}{\cellcolor[HTML]{DAE8FC}\textbf{Finish \%}} & \textbf{AVG Distance \%} \\ \hline
        \cellcolor[HTML]{FFFFFF}                                     & \textbf{No Noise}                                        & \multicolumn{1}{l|}{\cellcolor[HTML]{FFFFFF}0}          & \multicolumn{1}{l|}{\cellcolor[HTML]{FFFFFF}2}          & \multicolumn{1}{l|}{\cellcolor[HTML]{FFFFFF}0}          & \multicolumn{1}{l|}{\cellcolor[HTML]{FFFFFF}2.22}               & 32.22                      & \multicolumn{1}{l|}{\cellcolor[HTML]{FFFFFF}0}          & \multicolumn{1}{l|}{\cellcolor[HTML]{FFFFFF}1}          & \multicolumn{1}{l|}{\cellcolor[HTML]{FFFFFF}2}          & \multicolumn{1}{l|}{\cellcolor[HTML]{FFFFFF}3.33}               & 36.66                      & \multicolumn{1}{l|}{\cellcolor[HTML]{DAE8FC}2.78}               & 34.44                      \\ \cline{2-14} 
        \multirow{-2}{*}{\cellcolor[HTML]{FFFFFF}\textbf{DQN}}       & \textbf{Noise}                                           & \multicolumn{1}{l|}{\cellcolor[HTML]{FFFFFF}0}          & \multicolumn{1}{l|}{\cellcolor[HTML]{FFFFFF}0}          & \multicolumn{1}{l|}{\cellcolor[HTML]{FFFFFF}1}          & \multicolumn{1}{l|}{\cellcolor[HTML]{FFFFFF}1.11}               & 30.64                      & \multicolumn{1}{l|}{\cellcolor[HTML]{FFFFFF}2}          & \multicolumn{1}{l|}{\cellcolor[HTML]{FFFFFF}1}          & \multicolumn{1}{l|}{\cellcolor[HTML]{FFFFFF}3}          & \multicolumn{1}{l|}{\cellcolor[HTML]{FFFFFF}6.67}               & 35.10                      & \multicolumn{1}{l|}{\cellcolor[HTML]{DAE8FC}3.89}               & 32.87                      \\ \hline
        \cellcolor[HTML]{FFFFFF}                                     & \textbf{No Noise}                                        & \multicolumn{1}{l|}{\cellcolor[HTML]{FFFFFF}2}          & \multicolumn{1}{l|}{\cellcolor[HTML]{FFFFFF}4}          & \multicolumn{1}{l|}{\cellcolor[HTML]{FFFFFF}5}          & \multicolumn{1}{l|}{\cellcolor[HTML]{FFFFFF}12.22}              & 42.61                      & \multicolumn{1}{l|}{\cellcolor[HTML]{FFFFFF}6}          & \multicolumn{1}{l|}{\cellcolor[HTML]{FFFFFF}3}          & \multicolumn{1}{l|}{\cellcolor[HTML]{FFFFFF}2}          & \multicolumn{1}{l|}{\cellcolor[HTML]{FFFFFF}12.22}              & 31.15                      & \multicolumn{1}{l|}{\cellcolor[HTML]{DAE8FC}12.22}              & 36.88                      \\ \cline{2-14} 
        \multirow{-2}{*}{\cellcolor[HTML]{FFFFFF}\textbf{DQN Noise}} & \textbf{Noise}                                           & \multicolumn{1}{l|}{\cellcolor[HTML]{FFFFFF}7}          & \multicolumn{1}{l|}{\cellcolor[HTML]{FFFFFF}5}          & \multicolumn{1}{l|}{\cellcolor[HTML]{FFFFFF}1}          & \multicolumn{1}{l|}{\cellcolor[HTML]{FFFFFF}14.44}              & 48.82                      & \multicolumn{1}{l|}{\cellcolor[HTML]{FFFFFF}1}          & \multicolumn{1}{l|}{\cellcolor[HTML]{FFFFFF}1}          & \multicolumn{1}{l|}{\cellcolor[HTML]{FFFFFF}2}          & \multicolumn{1}{l|}{\cellcolor[HTML]{FFFFFF}4.44}               & 25.71                      & \multicolumn{1}{l|}{\cellcolor[HTML]{DAE8FC}9.44}               & 37.27                      \\ \hline
        \cellcolor[HTML]{FFFFFF}                                     & \textbf{No Noise}                                        & \multicolumn{1}{l|}{\cellcolor[HTML]{FFFFFF}1}          & \multicolumn{1}{l|}{\cellcolor[HTML]{FFFFFF}0}          & \multicolumn{1}{l|}{\cellcolor[HTML]{FFFFFF}0}          & \multicolumn{1}{l|}{\cellcolor[HTML]{FFFFFF}1.11}               & 20.89                      & \multicolumn{1}{l|}{\cellcolor[HTML]{FFFFFF}18}         & \multicolumn{1}{l|}{\cellcolor[HTML]{FFFFFF}21}         & \multicolumn{1}{l|}{\cellcolor[HTML]{FFFFFF}17}         & \multicolumn{1}{l|}{\cellcolor[HTML]{FFFFFF}\textbf{62.22}}     & \textbf{79.88}             & \multicolumn{1}{l|}{\cellcolor[HTML]{DAE8FC}31.67}              & 50.39                      \\ \cline{2-14} 
        \multirow{-2}{*}{\cellcolor[HTML]{FFFFFF}\textbf{TD3}}       & \textbf{Noise}                                           & \multicolumn{1}{l|}{\cellcolor[HTML]{FFFFFF}0}          & \multicolumn{1}{l|}{\cellcolor[HTML]{FFFFFF}0}          & \multicolumn{1}{l|}{\cellcolor[HTML]{FFFFFF}1}          & \multicolumn{1}{l|}{\cellcolor[HTML]{FFFFFF}1.11}               & 18.68                      & \multicolumn{1}{l|}{\cellcolor[HTML]{FFFFFF}14}         & \multicolumn{1}{l|}{\cellcolor[HTML]{FFFFFF}21}         & \multicolumn{1}{l|}{\cellcolor[HTML]{FFFFFF}16}         & \multicolumn{1}{l|}{\cellcolor[HTML]{FFFFFF}56.67}              & 76.06                      & \multicolumn{1}{l|}{\cellcolor[HTML]{DAE8FC}28.89}              & 47.37                      \\ \hline
        \cellcolor[HTML]{FFFFFF}                                     & \textbf{No Noise}                                        & \multicolumn{1}{l|}{\cellcolor[HTML]{FFFFFF}23}         & \multicolumn{1}{l|}{\cellcolor[HTML]{FFFFFF}23}         & \multicolumn{1}{l|}{\cellcolor[HTML]{FFFFFF}27}         & \multicolumn{1}{l|}{\cellcolor[HTML]{FFFFFF}\textbf{81.11}}     & \textbf{90.62}             & \multicolumn{1}{l|}{\cellcolor[HTML]{FFFFFF}1}          & \multicolumn{1}{l|}{\cellcolor[HTML]{FFFFFF}0}          & \multicolumn{1}{l|}{\cellcolor[HTML]{FFFFFF}0}          & \multicolumn{1}{l|}{\cellcolor[HTML]{FFFFFF}1.11}               & 13.35                      & \multicolumn{1}{l|}{\cellcolor[HTML]{DAE8FC}\textbf{41.11}}     & \textbf{51.99}             \\ \cline{2-14} 
        \multirow{-2}{*}{\cellcolor[HTML]{FFFFFF}\textbf{TD3 Noise}} & \textbf{Noise}                                           & \multicolumn{1}{l|}{\cellcolor[HTML]{FFFFFF}20}         & \multicolumn{1}{l|}{\cellcolor[HTML]{FFFFFF}20}         & \multicolumn{1}{l|}{\cellcolor[HTML]{FFFFFF}26}         & \multicolumn{1}{l|}{\cellcolor[HTML]{FFFFFF}73.33}              & 83.76                      & \multicolumn{1}{l|}{\cellcolor[HTML]{FFFFFF}0}          & \multicolumn{1}{l|}{\cellcolor[HTML]{FFFFFF}1}          & \multicolumn{1}{l|}{\cellcolor[HTML]{FFFFFF}0}          & \multicolumn{1}{l|}{\cellcolor[HTML]{FFFFFF}1.11}               & 12.88                      & \multicolumn{1}{l|}{\cellcolor[HTML]{DAE8FC}37.22}              & 48.32                      \\ \hline
        \end{tabular}
        \end{adjustbox}
        \label{tab:oval-track-results}
        \end{table*}

\section{Discussion}
    % Discuss the results and performance - learning, limitations, and suggested improvements
    The success of training in simulation to the real world using the cone/Aruco state representation is promising for future development, as scaling to a real-world race car will further increase the challenges of training on the real-world platform. However, several limitations and issues still need to be investigated further. 
    
    \subsection{Rotational Control Jitter}
        One key observation was a high level of jitter in the models' turning. This is an expected behaviour as the reward function is solely based on the angle to the centre-point of two cones and not the rate of change of its actions. Introducing a more complex reward function that evaluates the smoothness of rotational control could see a drastic reduction in the jitter.
        
   \subsection{Evaluation of Models}    
        The testing shows that both TD3 models consistently outperform the DQN models. The TD3 model trained without noise has the best performance overall in segment testing and real-world testing. The noisy TD3 model only slightly outperforms the noiseless one in the oval track testing. The results show that the TD3 models are better suited to the task of controlling a racing robot, and the noiseless model appears to be the best overall. 
        
        Adding noise during training appears to have no significant or consistent effect on performance. It was unexpected for the noiseless TD3 to outperform the noisy TD3 as, in theory, the noisy model should have been more adaptable and robust. However, this could probably be explained by the relatively small magnitude of the noise applied in our research. Therefore the noiseless model was able to navigate without a problem. It is possible that increasing the magnitude of noise to more accurately represent human errors associated with setting up a real track may result in the noisy TD3 model being the best.
    
    \subsection{TD3 Turning Bias}
        Both TD3 models learnt behaviours that allowed them to navigate well in one direction but not in the other. When averaging the results for each model on both noisy and noiseless tracks, the TD3 model trained without noise completed the track 59.45\% (107/180) of the time when turning left (clockwise), and the noisy TD3 model completed the track 77.22\% (139/180) of the time when turning right (anti-clockwise). Yet, both models only completed the track 1.11\% (2/180) while going in the opposite direction. This biased turning behaviour was not observed in real-world testing, which may be attributed to the fact that the real-world testing was done on a single-track segment due to space limitations.
        
        As both TD3 models developed a similar behaviour, albeit in different directions, the models could be overfitting to one direction based on a potential bias in the randomised training track segments (i.e., being presented with slightly more rights than lefts in the exploration). Training the models using pseudo-random training segments (i.e., giving the same number of lefts and rights) and increasing the number of exploration episodes could help test this theory and overcome the issue. Another explanation could be that the TD3 network is overly sensitive to its hyperparameters. A parameter sweep to find optimal parameters for the problem could help produce a more reliable model.
    
    \subsection{Centre-line Deviation}
        Another interesting finding showed that \textbf{all} test episodes with an average deviation from the centre-line $> 0.5$ units had track completions of $< 50\%$. By extension, all episodes in which the model completed the track had an average deviation from the centre-line was $\leq 0.5$ units. These results may be due to an episode being terminated if the robot approaches too close to a marker to keep the robot from going off track. An alternative method of detecting the robot’s position relative to the overall track rather than individual markers could be used to detect when the robot has left the track instead. This would allow the robot to deviate further from the centre-line (get closer to lane markers) without prematurely ending an episode. Increasing the number of steps (currently 10) the robot can take without seeing a “pair” before terminating the episode might also allow the robot some more time to correct its trajectory and realign with the track.
    
    \subsection{Sim-to-real Transfer Issues: Camera}
        Although the final autonomous FSAE vehicle should be able to operate in many/any lighting conditions, the current setup using the Realsense D435 camera showed issues with reliable Aruco marker detection in varying lighting conditions. During real-world testing, we found that the Realsense D435 camera struggled to accurately and consistently detect ArUco markers in low-lit and over-lit conditions, even indoors. The camera only performed acceptably within a small range of lighting conditions. This could partially be attributed to the quality of the camera but also potentially contributed to by the black and white colour scheme of the ArUco markers, making shadows (dark/black) and reflections (bright/white) on the markers affect detection by “altering” the shapes. Changing weather and sunlight outside caused varying levels of natural light in the room, even on the same day. As such, the detection accuracy was poorer than the relatively reliable detection in simulation. 
        
        Motion blur was also a significant problem when testing the robot in the real world with the same linear and angular speeds defined in the simulation. During sharper turns at these speeds, the camera showed a lot of motion blur and, as a result, constantly detected no ArUco markers causing the robot to move unexpectedly or terminate the episode. The linear and angular speeds had to be scaled down by 50\% compared to simulation speeds to allow the camera to detect the markers reliably. These detection issues could be overcome simply by using a higher quality camera with a higher frame rate and dynamic range. Replacing the ArUco markers with brightly coloured cones as used in the FSAE Germany DV events with an effective detection algorithm may also improve the detection of the lane markings. Making both of these changes would ensure the robotic system is robust and capable of navigating various outdoor environments.
    
    \subsection{Sim-to-real Transfer Issues: Space}
        In the real world, all the models did the worst on the right turn segment, regardless of their behaviours observed in the simulation. It should be noted that these real-world tests were performed on single-track segments and may show significantly different behaviours in multi-segment sections of track. Another critical point to note is that although the right turn segment was set up in the same place as the straight and the left turn, it was placed facing the opposite direction due to limited space. This change in direction may have consistently hindered the performance of the models due to the different lighting conditions adding additional noise to the state.
    
    \subsection{Sim-to-real Transfer Issues: Noise}
        Another key limitation faced when testing the models in the real world related to the accuracy of the marker placements. The models were trained with a fixed turn radius and programmatically calculated marker spacing for all turn segments. Although a small degree of noise was added, the models required accurately spaced markers to behave as expected in the real world. The required marker placement and spacing accuracy was achieved using measuring tapes and protractors. Although some care will be taken to ensure standardised track specifications in FSAE DV events, the precision required by the current models is unreasonably precise. As such, the models would likely need to be trained with a higher magnitude of noise and varying turn radii. These alterations to the training process should improve the reliability of the models in navigating and completing a wide variety of tracks in both simulation and reality. Normalising the state space input and taking the relative difference rather than the actual position values could make the system robust to the specific spacing of the cones. The model could then make decisions based on the robot's position relative to the two markers, regardless of their separation.

    \section{Conclusions and Future Work}
        In this proposal, we have presented the preliminary work toward using RL for mobile robotic navigation of an FSAE race car on a Turtlebot2 robotic platform. We benchmarked two state-of-the-art RL algorithms in both continuous and discrete action spaces. The results show that the continuous action space algorithm, TD3, is better suited to navigate and control a robot on a race track than DQN. The successful transfer of training in simulation to the real world is promising for extending the work towards a full-scale FSAE racecar. As an FSAE vehicle would not use a differential drive, the action space would need to be adjusted. Still, the conceptual findings of our research should transfer across without significant issues. Our findings highlighted practical issues that need to be accounted for, including variations in marker placement, motion blur, and lighting. Suggestions for improvement have been given to expand upon our research. 
    
        Future work will include using cones instead of markers and seek to minimise the differences between simulation and the real world through improved visual systems. 
        Due to the robot's highly "jittery" motion learnt from the current actions and reward function, we recommend increasing the complexity of the action space and implementing a more sophisticated reward function to learn racing lines and reward smoother control. This research will then be applied to the navigation and control of an F1TENTH\footnote{\href{https://f1tenth.org/}{https://f1tenth.org/}} scale model race car.

% \section*{Acknowledgements}
% Acknowledgements

\bibliography{publications}

\begin{thebibliography}{}

\bibitem[\protect\citeauthoryear{Arulkumaran \bgroup \em et al.\egroup
  }{2017a}]{arulkumaran2017brief}
Kai Arulkumaran, Marc~Peter Deisenroth, Miles Brundage, and Anil~Anthony
  Bharath.
\newblock A brief survey of deep reinforcement learning.
\newblock {\em arXiv preprint arXiv:1708.05866}, 2017.

\bibitem[\protect\citeauthoryear{Arulkumaran \bgroup \em et al.\egroup
  }{2017b}]{arulkumaran2017deep}
Kai Arulkumaran, Marc~Peter Deisenroth, Miles Brundage, and Anil~Anthony
  Bharath.
\newblock Deep reinforcement learning: A brief survey.
\newblock {\em IEEE Signal Processing Magazine}, 34(6):26--38, 2017.

\bibitem[\protect\citeauthoryear{Bulog \bgroup \em et al.\egroup
  }{2019}]{Williams2019}
E.~Bulog, M.~Frost, and H.~Williams.
\newblock Deep reinforcement learning for path planning.
\newblock {\em Australasian Conference on Robotics and Automation, ACRA},
  2019-December, 2019.

\bibitem[\protect\citeauthoryear{{\c{C}}al{\i}{\c{s}}{\i}r and
  Pehlivano{\u{g}}lu}{2019}]{ccalicsir2019model}
Sinan {\c{C}}al{\i}{\c{s}}{\i}r and Meltem~Kurt Pehlivano{\u{g}}lu.
\newblock Model-free reinforcement learning algorithms: A survey.
\newblock In {\em 2019 27th Signal Processing and Communications Applications
  Conference (SIU)}, pages 1--4. IEEE, 2019.

\bibitem[\protect\citeauthoryear{Dewanto \bgroup \em et al.\egroup
  }{2020}]{dewanto2020average}
Vektor Dewanto, George Dunn, Ali Eshragh, Marcus Gallagher, and Fred Roosta.
\newblock Average-reward model-free reinforcement learning: a systematic review
  and literature mapping.
\newblock {\em arXiv preprint arXiv:2010.08920}, 2020.

\bibitem[\protect\citeauthoryear{FSG}{2022a}]{avhandbook2022}
FSG.
\newblock Fsg: Competition handbook 2022.
\newblock
  \url{https://www.formulastudent.de/fileadmin/user_upload/all/2022/rules/FSG22_Competition_Handbook_v1.1.pdf},
  2022.
\newblock Accessed: 2022-08-26.

\bibitem[\protect\citeauthoryear{FSG}{2022b}]{avrules2022}
FSG.
\newblock Fsg: Rules and documents.
\newblock \url{https://www.formulastudent.de/fsg/rules/}, 2022.
\newblock Accessed: 2022-08-26.

\bibitem[\protect\citeauthoryear{Fujimoto \bgroup \em et al.\egroup
  }{2018}]{fujimoto2018addressing}
Scott Fujimoto, Herke Hoof, and David Meger.
\newblock Addressing function approximation error in actor-critic methods.
\newblock In {\em International conference on machine learning}, pages
  1587--1596. PMLR, 2018.

\bibitem[\protect\citeauthoryear{Garrido-Jurado \bgroup \em et al.\egroup
  }{2014}]{garrido2014automatic}
Sergio Garrido-Jurado, Rafael Mu{\~n}oz-Salinas, Francisco~Jos{\'e}
  Madrid-Cuevas, and Manuel~Jes{\'u}s Mar{\'\i}n-Jim{\'e}nez.
\newblock Automatic generation and detection of highly reliable fiducial
  markers under occlusion.
\newblock {\em Pattern Recognition}, 47(6):2280--2292, 2014.

\bibitem[\protect\citeauthoryear{Gul \bgroup \em et al.\egroup
  }{2019}]{gul2019comprehensive}
Faiza Gul, Wan Rahiman, and Syed~Sahal Nazli~Alhady.
\newblock A comprehensive study for robot navigation techniques.
\newblock {\em Cogent Engineering}, 6(1):1632046, 2019.

\bibitem[\protect\citeauthoryear{Jun \bgroup \em et al.\egroup
  }{2018}]{jun2018autonomous}
NI~Jun, HU~Jibin, et~al.
\newblock Autonomous driving system design for formula student driverless
  racecar.
\newblock In {\em 2018 IEEE Intelligent Vehicles Symposium (IV)}, pages 1--6.
  IEEE, 2018.

\bibitem[\protect\citeauthoryear{Kiran \bgroup \em et al.\egroup
  }{2021}]{kiran2021deep}
B~Ravi Kiran, Ibrahim Sobh, Victor Talpaert, Patrick Mannion, Ahmad~A
  Al~Sallab, Senthil Yogamani, and Patrick P{\'e}rez.
\newblock Deep reinforcement learning for autonomous driving: A survey.
\newblock {\em IEEE Transactions on Intelligent Transportation Systems}, 2021.

\bibitem[\protect\citeauthoryear{Li \bgroup \em et al.\egroup
  }{2021}]{li2021behavior}
Juncheng Li, Maopeng Ran, Han Wang, and Lihua Xie.
\newblock A behavior-based mobile robot navigation method with deep
  reinforcement learning.
\newblock {\em Unmanned Systems}, 9(03):201--209, 2021.

\bibitem[\protect\citeauthoryear{Lillicrap \bgroup \em et al.\egroup
  }{2015}]{lillicrap2015continuous}
Timothy~P Lillicrap, Jonathan~J Hunt, Alexander Pritzel, Nicolas Heess, Tom
  Erez, Yuval Tassa, David Silver, and Daan Wierstra.
\newblock Continuous control with deep reinforcement learning.
\newblock {\em arXiv preprint arXiv:1509.02971}, 2015.

\bibitem[\protect\citeauthoryear{Liu \bgroup \em et al.\egroup
  }{2021}]{liu2021policy}
Yongshuai Liu, Avishai Halev, and Xin Liu.
\newblock Policy learning with constraints in model-free reinforcement
  learning: A survey.
\newblock In {\em Proceedings of the Thirtieth International Joint Conference
  on Artificial Intelligence}, 2021.

\bibitem[\protect\citeauthoryear{Masetty \bgroup \em et al.\egroup
  }{2021}]{masettycerebellum}
Bharath Masetty, Reuth Mirsky, Ashish Deshpande, Michael Mauk, and Peter Stone.
\newblock Is the cerebellum a model-based reinforcement learning agent?
\newblock {\em 20th International Conference on Autonomous Agents and
  Multiagent Systems}, 2021.

\bibitem[\protect\citeauthoryear{Mnih \bgroup \em et al.\egroup
  }{2013}]{mnih2013playing}
Volodymyr Mnih, Koray Kavukcuoglu, David Silver, Alex Graves, Ioannis
  Antonoglou, Daan Wierstra, and Martin Riedmiller.
\newblock Playing atari with deep reinforcement learning.
\newblock {\em arXiv preprint arXiv:1312.5602}, 2013.

\bibitem[\protect\citeauthoryear{Mohanan and
  Salgoankar}{2018}]{mohanan2018survey}
MG~Mohanan and Ambuja Salgoankar.
\newblock A survey of robotic motion planning in dynamic environments.
\newblock {\em Robotics and Autonomous Systems}, 100:171--185, 2018.

\bibitem[\protect\citeauthoryear{Niroui \bgroup \em et al.\egroup
  }{2019}]{niroui2019deep}
Farzad Niroui, Kaicheng Zhang, Zendai Kashino, and Goldie Nejat.
\newblock Deep reinforcement learning robot for search and rescue applications:
  Exploration in unknown cluttered environments.
\newblock {\em IEEE Robotics and Automation Letters}, 4(2):610--617, 2019.

\bibitem[\protect\citeauthoryear{Ram{\'\i}rez \bgroup \em et al.\egroup
  }{2021}]{ramirez2021model}
Jorge Ram{\'\i}rez, Wen Yu, and Adolfo Perrusqu{\'\i}a.
\newblock Model-free reinforcement learning from expert demonstrations: a
  survey.
\newblock {\em Artificial Intelligence Review}, pages 1--29, 2021.

\bibitem[\protect\citeauthoryear{Remonda \bgroup \em et al.\egroup
  }{2021}]{remonda2021formula}
Adrian Remonda, Sarah Krebs, Eduardo Veas, Granit Luzhnica, and Roman Kern.
\newblock Formula rl: Deep reinforcement learning for autonomous racing using
  telemetry data.
\newblock {\em arXiv preprint arXiv:2104.11106}, 2021.

\bibitem[\protect\citeauthoryear{Sato}{2019}]{sato2019model}
Yoshiharu Sato.
\newblock Model-free reinforcement learning for financial portfolios: a brief
  survey.
\newblock {\em arXiv preprint arXiv:1904.04973}, 2019.

\bibitem[\protect\citeauthoryear{Schulman \bgroup \em et al.\egroup
  }{2017}]{schulman2017proximal}
John Schulman, Filip Wolski, Prafulla Dhariwal, Alec Radford, and Oleg Klimov.
\newblock Proximal policy optimization algorithms.
\newblock {\em arXiv preprint arXiv:1707.06347}, 2017.

\bibitem[\protect\citeauthoryear{Silver \bgroup \em et al.\egroup
  }{2016}]{silver2016mastering}
David Silver, Aja Huang, Chris~J Maddison, Arthur Guez, Laurent Sifre, George
  Van Den~Driessche, Julian Schrittwieser, Ioannis Antonoglou, Veda
  Panneershelvam, Marc Lanctot, et~al.
\newblock Mastering the game of go with deep neural networks and tree search.
\newblock {\em nature}, 529(7587):484--489, 2016.

\bibitem[\protect\citeauthoryear{Sutton and
  Barto}{2018}]{sutton2018reinforcement}
Richard~S Sutton and Andrew~G Barto.
\newblock {\em Reinforcement learning: An introduction}.
\newblock MIT press, 2018.

\bibitem[\protect\citeauthoryear{Van~Hasselt \bgroup \em et al.\egroup
  }{2016}]{van2016deep}
Hado Van~Hasselt, Arthur Guez, and David Silver.
\newblock Deep reinforcement learning with double q-learning.
\newblock In {\em Proceedings of the AAAI conference on artificial
  intelligence}, 2016.

\bibitem[\protect\citeauthoryear{Wang \bgroup \em et al.\egroup
  }{2016}]{wang2016dueling}
Ziyu Wang, Tom Schaul, Matteo Hessel, Hado Hasselt, Marc Lanctot, and Nando
  Freitas.
\newblock Dueling network architectures for deep reinforcement learning.
\newblock In {\em International conference on machine learning}, pages
  1995--2003. PMLR, 2016.

\bibitem[\protect\citeauthoryear{Xiao \bgroup \em et al.\egroup
  }{2020}]{xiao2020motion}
Xuesu Xiao, Bo~Liu, Garrett Warnell, and Peter Stone.
\newblock Motion control for mobile robot navigation using machine learning: a
  survey.
\newblock {\em arXiv preprint arXiv:2011.13112}, 2020.

\bibitem[\protect\citeauthoryear{Xiao \bgroup \em et al.\egroup
  }{2022}]{xiao2022motion}
Xuesu Xiao, Bo~Liu, Garrett Warnell, and Peter Stone.
\newblock Motion planning and control for mobile robot navigation using machine
  learning: a survey.
\newblock {\em Autonomous Robots}, pages 1--29, 2022.

\bibitem[\protect\citeauthoryear{Zadok \bgroup \em et al.\egroup
  }{2019}]{zadok2019explorations}
Dean Zadok, Tom Hirshberg, Amir Biran, Kira Radinsky, and Ashish Kapoor.
\newblock Explorations and lessons learned in building an autonomous formula
  sae car from simulations.
\newblock {\em arXiv preprint arXiv:1905.05940}, 2019.

\bibitem[\protect\citeauthoryear{Zhang \bgroup \em et al.\egroup
  }{2018}]{zhang2018robot}
Kaichena Zhang, Farzad Niroui, Maurizio Ficocelli, and Goldie Nejat.
\newblock Robot navigation of environments with unknown rough terrain using
  deep reinforcement learning.
\newblock In {\em 2018 IEEE International Symposium on Safety, Security, and
  Rescue Robotics (SSRR)}, pages 1--7. IEEE, 2018.

\bibitem[\protect\citeauthoryear{Zhao \bgroup \em et al.\egroup
  }{2020}]{zhao2020sim}
Wenshuai Zhao, Jorge~Pe{\~n}a Queralta, and Tomi Westerlund.
\newblock Sim-to-real transfer in deep reinforcement learning for robotics: a
  survey.
\newblock In {\em 2020 IEEE Symposium Series on Computational Intelligence
  (SSCI)}, pages 737--744. IEEE, 2020.

\bibitem[\protect\citeauthoryear{Zhu and Zhang}{2021}]{zhu2021deep}
Kai Zhu and Tao Zhang.
\newblock Deep reinforcement learning based mobile robot navigation: A review.
\newblock {\em Tsinghua Science and Technology}, 26(5):674--691, 2021.

\end{thebibliography}
\bibliographystyle{named}
\end{document}